\documentclass{article}
\usepackage{arxiv}
\usepackage{amsmath}
\usepackage[utf8]{inputenc} 
\usepackage[T1]{fontenc}    
\usepackage{hyperref}       
\usepackage{url}            
\usepackage{booktabs}       
\usepackage{amsfonts}       
\usepackage{nicefrac}       
\usepackage{microtype}      
\usepackage{lipsum}
\usepackage{graphicx}
\graphicspath{ {./images/} }
\usepackage[authoryear]{natbib}
\RequirePackage{graphicx}
\usepackage{booktabs}
\usepackage{tikz} 
\usepackage{verbatim}
\usepackage{caption}
\usepackage{subcaption}
\usepackage{bbm}
\usepackage{mathabx}

\title{Ecotransformer: Attention without Multiplication}

\author{
 Xin Gao \\
  Department of Mathematics and Statistics\\
  York University\\
  Toronto, M3J 1P3 \\
  \texttt{xingao@yorku.ca} \\
   \And
 Xingming Xu \\
  Department of Computer Science\\
  University of California Davis\\
  Davis, CA 95616 \\
  \texttt{xmxu@ucdavis.edu} 
   \AND
   Shirin Amiraslani \\
  Department of Mathematics and Statistics\\
  York University\\
  Toronto, M3J 1P3 \\
  \texttt{shirinamiraslani@gmail.com} 
   \And
   Hong Xu \\
   Market Risk Model  \\
  Bank of Montreal Financial Group\\
  \texttt{jiaxuhongwh@gmail.com} \\
}

\begin{document}
\maketitle
\begin{abstract}
The Transformer, with its scaled dot-product attention mechanism, has become a foundational architecture in modern AI. However, this mechanism is computationally intensive and incurs substantial energy costs. We propose a new Transformer architecture \textbf{\textit{EcoTransformer}}, in which the output context vector is constructed as the convolution of the values using a Laplacian kernel, where the distances are measured by the $L_1$ metric between the queries and keys. 
Compared to dot-product based attention, the new attention score calculation is free of matrix multiplication. 
It performs on par with, or even surpasses, scaled dot-product attention in NLP, bioinformatics, and vision tasks, while consuming significantly less energy.
\end{abstract}

\thispagestyle{fancy}
\fancyfoot[C]{\small \textcopyright~2025 Xin Gao.
Version 2 of this manuscript supersedes version 1. The authors have updated the license to restrict commercial use. This work is licensed for academic and non-commercial use only. Certain aspects of this work are subject to patent protection. 
}

\section{Introduction}

The ability to model dependencies in sequences and structured data lies at the heart of many machine learning applications. Traditional deep learning architectures such as recurrent neural networks (RNNs) and convolutional neural networks (CNNs) struggled with long-range dependencies and fixed-size receptive fields. Attention mechanisms emerged as a powerful paradigm shift, allowing models to dynamically assign importance to various input elements, thereby addressing limitations in prior architectures.  \cite{bahdanau2014neural} introduced additive attention, where alignment scores between queries and keys are learned via a feedforward network. This pioneering work laid the groundwork for modern architectures. \cite{luong2015effective} proposed multiplicative attention, simplifying computation by using dot-products between hidden states. The Transformer architecture by \cite{vaswani2017attention} formalized the concept of scaled dot-product attention, computing relevance scores as softmax-normalized dot products scaled by the dimension of the feature space. Crucially, the Transformer introduced multi-head attention, enabling models to capture diverse subspace relationships by learning multiple attention distributions in parallel. 
Since their introduction in neural machine translation, attention mechanisms have become foundational in Transformer architecture, redefining state-of-the-art performance across NLP, vision, and beyond. 

While effective, traditional attention mechanisms incur quadratic complexity with respect to sequence length. Numerous innovations have emerged to address this bottleneck. Sparse attention restricts attention computations to predefined patterns, as seen in models like Longformer \citep{beltagy2020longformer}, BigBird \citep{zaheer2020big}, and Sparse Transformer \citep{child2019generating}. These methods maintain expressivity while reducing complexity. Low-rank and projection-based techniques, including Linformer \citep{wang2020linformer} and Nyströmformer \citep{xin2022nystromformer}, approximate the full attention matrix using low-rank projections or sampling, enabling linear time complexity. Kernel-based methods, such as Performer \citep{choromanski2021rethinking}, apply random feature maps to approximate the softmax kernel used in attention, using the FAVOR+ algorithm to retain theoretical guarantees while achieving linear efficiency. Convolution-based approaches like Hyena \citep{poli2023hyena} replace attention entirely with deep implicit convolutions, offering scalable modeling for long sequences. Alternative paradigms like Synthesizer \citep{tay2021synthesizer} and Mamba \citep{gu2023mamba} eliminate explicit key-query interactions, using learned patterns or recurrent state spaces to encode dependencies. Recent memory-efficient implementations, including FlashAttention \citep{dao2022flashattention, dao2023flashattention} and xFormers \citep{xformers2022}, focus on low-level optimizations and hardware-friendly execution without changing the attention formulation itself.

Despite the remarkable success of Transformer models across a broad range of AI applications, the standard dot-product attention mechanism remains a major computational bottleneck. The core cost arises from the $QK^T$ matrix multiplication, which scales as $
O(N^2D_k)$, where $N$ is the sequence length and $D_k$ is the hidden dimension. For example, in GPT-3, with a sequence length of 2048 and a hidden size of 12,288, each token generation involves attention computations across 96 heads and 96 layers, resulting in billions of floating-point operations per forward pass. This computational intensity translates into substantial energy demands. In early 2023, ChatGPT was estimated to consume approximately 564 megawatt-hours per day, equivalent to the daily electricity usage of about 18,000 U.S. households \citep{luo2024addition}. In more extreme cases, the annual energy consumption of Google’s AI services has been estimated at 29.3 terawatt-hours, comparable to the electricity consumption of an entire country such as Ireland \citep{de2023growing}. This motivates the development of more efficient alternatives to dot-product attention that can achieve comparable expressiveness with reduced computational overhead. In this manuscript, we introduce a novel method of attention score calculation that entirely eliminates the need for matrix multiplication, departing from the traditional dot-product formulation at the core of standard Transformer architectures. 

Our motivation stems from the observation that the attention mechanism fundamentally operates by constructing the output context vector as a weighted sum of the value vectors. The weights are determined by the dot-product between the query and key vectors—higher dot-products indicate stronger relevance and thus receive greater attention. 
Although the dot product is a highly effective mathematical tool for measuring the relevance between two variables, it is unlikely that the human brain relies on dot-product computations to assess relationships. Instead, humans tend to understand relatedness by imagining objects in space and judging their distance. Our brains are naturally attuned to recognizing whether things are close together or far apart, rather than performing vector multiplications. In this intuitive spatial framework, short distances signify strong relationships, while long distances suggest weak or no connection. Thus, measuring dependency or relatedness can be reinterpreted as measuring distance, which aligns closer with how the human brain processes information. Inspired by the human brain’s tendency to assess relevance through spatial proximity, we propose the EcoTransformer attention mechanism that leverages the functional relationship between distance and dot-product. We construct the output context vector as the convolution of the values using distance kernel, where the distances are measured between queries and keys instead of dot-product between them. Because the new architecture subsumes scaled dot-product attention as a special case, the majority of existing techniques for reducing computational complexity to sub-quadratic or linear levels can still be applied. By avoiding expensive matrix multiplication in attention score calculation, our approach significantly reduces energy consumption, making it particularly advantageous for resource-constrained environments. 
Our findings show that with proper tuning, this alternative attention mechanism matches and rivals the performance of dot-product attention across diverse domains, reinforcing its viability as a practical and efficient replacement.

\section{Distance-Based Attention Score }
In the Transformer model, the scaled dot-product attention mechanism is defined as follows. Given an input sequence embedding \( X \), which includes positional encodings, we compute the query, key, and value matrices as:
\[
Q = XW^{(Q)}, \quad K = XW^{(K)}, \quad V = XW^{(V)},
\]
where \( W^{(Q)} \), \( W^{(K)} \), and \( W^{(V)} \) are learnable weight matrices. The attention score matrix \( \alpha \in \mathbb{R}^{N \times N} \) is then calculated using:
\[
\alpha = \text{softmax}\left( \frac{QK^T}{\sqrt{D_k}} \right),
\]
and the output of the attention layer is:
\[
O = \alpha V.
\]
Each output vector \( \mathbf{O}_i \) is a weighted sum of the value vectors:
\[
\mathbf{O}_i = \sum_{j=1}^N \alpha_{ij} \mathbf{V}_j,
\]
where $$\alpha_{ij}=\frac{e^{<\mathbf{Q}_i, \mathbf{K}_j>/\sqrt{D_k}}}{\sum_{j'} e^{<\mathbf{Q}_i, \mathbf{K}_{j'}>/\sqrt{D_k}}},$$ and $\mathbf{Q}_i, \mathbf{K}_j, \mathbf{V}_j$ are the rows of the query, key and value matrices, respectively. The dot product $<\mathbf{Q}_i, \mathbf{K}_j>$ can be reformulated in terms of $L_2$ distance:
$$
<\mathbf{Q}_i, \mathbf{K}_j>=\frac 1 2\{||\mathbf{Q}_i||_2^2+ ||\mathbf{K}_j||_2^2|-||\mathbf{Q}_i-\mathbf{K}_j||_2^2\}.
$$
If queries and keys are properly normalized with unit $L_2$ norms, then
$$
\alpha_{ij}=\frac{e^{-\frac {1} {2/\sqrt{D_k}}||\mathbf{Q}_i-\mathbf{K}_j||_2^2}}{\sum_{j'} e^{-\frac {1} {2/\sqrt{D_k}}||\mathbf{Q}_i-\mathbf{K}_{j'}||_2^2}}.
$$
This equation demonstrates that the attention weight is functionally dependent on the $L_2$ distances between the queries and keys.  
Thus, we build a connection between the attention mechanism and the distance metric. Our proposed distance-based attention mechanism is constructed as follows. Given $Q$ and $K$ matrices, we construct an operator $L$ such that $L(Q, K)$ generates a matrix $L,$ where 
$$
L_{ij}=-\text{distance}(\mathbf{Q}_i, \mathbf{K}_j).
$$
The $L$ matrix records all the pairwise negative distances between the queries and keys.
The new attention  $\alpha^{new}\in \mathcal{R}^{N\times N}$ is computed by 
\begin{equation}
\label{newalpha}
\alpha^{new}=\text{softmax}(\frac{\lambda L}{\sqrt{D_k}}),
\end{equation}
$$O=\alpha^{new} V,$$ where $\lambda$ is a tuning parameter.
The output context vectors are weighted sums of the value vectors defined
as follows:
\begin{equation}
\label{newout}   
\mathbf{O}_i=\sum_{j=1}^N \alpha^{new}_{ij} \mathbf{V}_j.
\end{equation}

Many distance measures can be applied to the attention mechanism including the general $L_p$ distance with $p\geq 1.$  The attention based on squared $L_2$-distance with $\lambda=\frac 1 2$ is equivalent to the scaled dot-product attention applied on normalized queries and keys. In this paper, we focus on the attention mechanism based on $L_1$ distance, where 
$$
L_{ij}=-|\mathbf{Q}_i-\mathbf{K}_j|_1=-\sum_{m=1}^{D_k} |Q_{im}-K_{jm}|.
$$

\section{The Convolution Mechanism}

As \cite{vaswani2017attention} stated ``Attention is all you need," we can further elaborate: ``Attention depends on distance." 
Given a query $\mathbf{Q}$ and a key $\mathbf{K}$, let $\vec{\mathbf{D}}=\mathbf{Q}-\mathbf{K}$ denote the difference vector with $\vec{\mathbf{D}}=(d_1,\dots,d_{D_k}).$ To model the relationship between attention and distance, we introduce an attention function $\alpha(.)$ such that the attention weight
$\alpha(\vec{D})=c\prod_{m=1}^{D_k} k(d_m)$, where $c$ is a normalizing constant and
\[
k(d) =
\begin{cases}
\exp\{-\frac{d^2}{2\sqrt{D_k}}\}, & \text{for scaled dot-product attention}\,(L_2), \\
\exp\{-\frac{\lambda |d|}{\sqrt{D_k}}\}, & \text{for}\, L_1 \,\text{attention}.
\end{cases}
\]
Thus, the context vector from Equations (\ref{newalpha}) and (\ref{newout}) can be viewed as the convolution of the values using the distance kernel:
 \[
O_i=\sum_{j=1}^{N}\alpha^{new}_{ij} \mathbf{V}_j= \sum_{j=1}^{N}\alpha(\mathbf{Q}_i-\mathbf{K}_j) \mathbf{V}_j,
 \]
 where the attention function corresponds to the Gaussian kernel and Laplacian kernel for the scaled dot-product attention and $L_1$ attention, respectively. The tuning parameter $\lambda$ can be considered as the bandwidth of the kernel function. In comparison to the Hyena architecture \citep{poli2023hyena}, its convolution distance kernel refers to a data-independent kernel that encodes positional or temporal distances between tokens. It is not based on the values of the tokens themselves, but rather on their relative positions in the sequence. In scaled dot-product attention ad $L_1$ attention, however, the distances are measured between the tokens' projected embeddings - queries and keys.


\begin{figure}[h]
    \centering
    \includegraphics[width=0.7\textwidth]{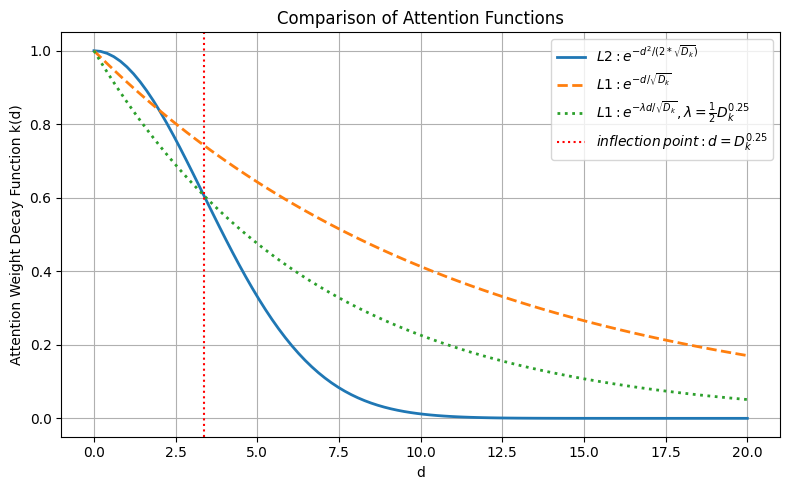}
    \caption{Comparison of Attention functions: $e^{-d^2/2\sqrt{D_k}}$ vs $e^{-\lambda |d|/\sqrt{D_k}}$}
    \label{fig:attention}
\end{figure}

We examine how the $L_2$ and $L_1$ distance metrics affect attention weights differently. Consider a single embedding dimension where the query and key differ by an absolute value of \( |d| \). Under the $L_2$-based attention mechanism, the attention weight is reduced by a multiplicative factor of \( e^{-d^2 / (2\sqrt{D_k})} \), whereas under the $L_1$-based mechanism, it is reduced by \( e^{-\lambda |d| / \sqrt{D_k}} \). In the exponential scale, the $L_1$ metric results in a linear decay with respect to \( |d| \), while the $L_2$ metric induces a quadratic decay. Figure~\ref{fig:attention} illustrates the decay functions \( k(d) \) for different attention mechanisms, plotted against the absolute difference \( |d| \). Compared to $L_1$ with $\lambda=1$, the $L_2$ function penalizes larger distances more severely, causing attention to focus on tokens that are very close in representation space, i.e., highly dependent tokens. In contrast, the $L_1$-based function distributes attention more broadly, allowing influence from tokens that are moderately or even weakly related. As the parameter \( \lambda \) increases, the $L_1$ decay curve approaches the shape of the $L_2$ curve. In particular, when \( \lambda = 1/2 D_k^{1/4} \), the $L_1$ and $L_2$ curves intersect at the inflection point of the $L_2$ curve. In this case, the two attention functions exhibit similar behaviors over a range of distances.

Determining the optimal attention function for assigning attention weights based on distance is ultimately task-dependent. It depends on the joint distribution of token representations and the structure of their dependencies. Importantly, stacking multiple attention layers allows the model to incorporate information not only from directly related tokens, but also from those that are indirectly connected. As each layer aggregates first-order dependencies, deeper layers progressively integrate second-order and higher-order relations. Therefore, the choice of attention decay function significantly influences how quickly information from various dependency levels is absorbed into the context vector. As shown in Figure~\ref{fig:attention}, by tuning the parameter \( \lambda \), the $L_1$-based mechanism defines a broad family of attention functions. Among these, one can select the optimal function by empirically identifying the value of \( \lambda \) that yields the best performance. In the numerical studies that follow, we demonstrate that with proper tuning, the $L_1$-based attention mechanism not only matches but often surpasses the performance of the standard scaled dot-product attention across a variety of tasks.

\section{Computational Complexity}

In comparison,  the $ij$th entry of $QK^T$ is $$<\mathbf{Q}_i, \mathbf{K}_j>=\sum_{m=1}^{D_k} Q_{im}*K_{jm},$$
whereas the $L_1$ distance matrix has entry
$$
L_{ij}=-|\mathbf{Q}_i-\mathbf{K}_j|_1=-\sum_{m=1}^{D_k} |Q_{im}-K_{jm}|.
$$
The computational complexity of the dot-product attention mechanism involves \(N^2 D_k\) multiplications and \(N^2 D_k\) additions. In contrast, the $L_1$-based attention mechanism replaces multiplications with \(N^2 D_k\) absolute difference operations and maintains the same number of additions. Although both methods have the same asymptotic complexity, the $L_1$-based approach offers a significant practical advantage by replacing multiplication operations with addition operations, which are substantially less expensive in computation.

On most modern hardware, additions are faster, consume less power, and incur lower latency than multiplications, especially at scale. According to \cite{cao2021irene}, attention mechanisms account for approximately 37.94\% of the total energy consumption in GPT-2. In the ChatGPT API, attention computation is estimated to consume 30--40\% of inference energy, making it one of the top energy drivers alongside feedforward layers, softmax, embeddings, and I/O operations. Energy efficiency differences between addition and multiplication operations are well documented. \cite{horowitz20141} reports that multiplying two 32-bit floating-point numbers (FP32) requires approximately 3.7 picojoules (pJ), compared to just 0.9 pJ for an addition---making multiplication 4.11 times more energy intensive. This suggests that replacing multiplications with additions in attention score computation can theoretically reduce energy usage by up to 61\% for this module. 
However, while the algorithm achieves theoretical gains in computational and energy efficiency, it is currently limited by hardware constraints. Existing GPU architectures, such as NVIDIA’s Tensor Cores, are heavily optimized for dense matrix multiplications and dot-product operations, giving traditional Transformer models a performance edge in practice. Addition-based algorithms do not \textit{yet} benefit from equivalent low-level hardware acceleration. To fully leverage the benefits of the proposed method, support from hardware manufacturers is essential. Future GPUs and AI accelerators would need to provide specialized instruction sets or dedicated units optimized for large-scale addition and absolute difference operations.

\section{L1 Attention Model with Linear Complexity}
The $L_1$ attention can be applied to other lower complexity attention architecture to reduce the complexity from $O(N^2D_k)$ to $O(ND_k).$  Consider the Longformer architecture, which can process very long sequences by sparsifying attention. The method computes local sliding window attention so that each token attends only to neighbors. It also computes global attention for selected important tokens. We propose the following $L_1$-Longformer structure as follows. Let $\text{Att}_t$ represent the attention weights for the tokens within the sliding neighborhood window of length $w$ and $\text{Att}_g$ represent the global attention weights for the selected token. The overall attention weights combines the weights from both sources:
\begin{align}
&\text{Att}_t=\text{softmax}(\frac{\lambda L(Q_t, K_{[t-\frac w 2:t+\frac w 2]})}{\sqrt{D_k}} V_{[t-\frac w 2:t+\frac w 2]}),\\
&\text{Att}_g=\text{softmax}(\frac{\lambda L(Q_g, K)}{\sqrt{D_k}} V)\\
&\text{output}_t=\text{Att}_t+\text{Att}_g.
\end{align}
The $L_1$ attention can also be integrated in the projection-based Linformer architecture as follows:
\[
\text{Attention}(Q,K,V)=\text{softmax}(\frac{\lambda L(Q, E_K K)}{\sqrt{D_k}} E_V V),
\]
where $E_K$ and $E_V$ are $k \times N$ projection matrices.

\section{Experimental Results}
\subsection{NLP applications}
To evaluate the effectiveness of the proposed $L_1$-based attention mechanism, we conduct experiments on a diverse set of natural language understanding benchmarks: SciQ, StoryCloze, HellaSwag and BoolQ. These datasets are widely used to assess reasoning and comprehension capabilities in AI models. SciQ is a multiple-choice science question-answering dataset based on middle school exam questions, requiring basic scientific knowledge and inference skills. StoryCloze evaluates narrative coherence by asking models to choose the most appropriate ending to a four-sentence story.  HellaSwag is a commonsense reasoning benchmark that tests a model’s ability to select the most plausible sentence continuation—often a challenge even for large-scale language models. BoolQ consists of yes/no questions derived from Wikipedia passages, requiring factual reasoning grounded in context. 

By evaluating across these varied tasks, we compare the performance of the proposed $L_1$ attention mechanism with that of the standard scaled dot-product attention, focusing on accuracy under different types of linguistic and logical reasoning demands. All models are trained under identical settings---batch size, embedding size, sequence length, number of heads, number of layers, learning rate, and number of epochs---to ensure a fair comparison. Table \ref{NLP} shows that the $L_1$-based method with \( \lambda = 1 \) achieves performance very close to that of the dot-product baseline. The observed accuracy difference ranges from \(-0.0018\) to \(-0.0110\), indicating minimal degradation. Moreover, when the optimal \( \lambda \) is selected via grid search, the $L_1$ method outperforms the dot-product method on three out of the four benchmarks. Although the optimal \( \lambda \) varies across tasks, setting \( \lambda = 3 \) yields robust performance across all datasets. In this configuration, the accuracy differences relative to the dot-product method are \( +0.0190 \), \( -0.0102 \), \( -0.0026 \), and \( +0.0034 \) on SciQ, StoryCloze, HellaSwag, and BoolQ, respectively. These results suggest that a properly tuned $L_1$ attention mechanism can achieve performance comparable to, and in some cases better than, the scaled dot-product attention across a range of NLP tasks. This highlights its potential as an efficient alternative without compromising accuracy.

\begin{table}[h!]
\centering
\caption{Accuracy Result of $L_1$ Attention and Dot-product Attention on NLP Datasets}
\label{NLP}
\begin{tabular}{lcccc}
\toprule
\textbf{Method}   &SciQ &StoryCloze &HellaSwag & BoolQ   \\
\midrule
dot-product   & 0.4020 & 0.6589 & \color{green}{0.2898} & 0.6376\\
$L_1$, $\lambda=1$ & 0.3910 & 0.6524          & 0.2894   & 0.6358\\
$L_1$, $\lambda=2$ & 0.3910 & 0.6471          & 0.2891  & 0.6413 \\
$L_1$, $\lambda=3$  & \color{green}{0.4210} & 0.6487 & 0.2872 & 0.6410\\
$L_1$, $\lambda=4$  & 0.3540 & 0.6400          & 0.2780 & 0.6446\\
$L_1$, $\lambda=5$  & 0.3300 & 0.6449            & 0.2791  & 0.6388\\
$L_1$, $\lambda=15$  & 0.3090 & \color{green}{0.7139}  & 0.2706  & \color{green}{0.6535}\\
\bottomrule
\end{tabular}
\end{table}

\subsection{Biological and Vision Applications}
To evaluate the generalizability and efficiency of the proposed transformer architecture, we conduct comprehensive comparative studies across three biologically diverse datasets: The Cancer Genome Atlas (TCGA), METABRIC, and a TCR--epitope classification dataset derived from VDJdb. TCGA and METABRIC are large-scale cancer genomics cohorts that provide transcriptomic profiles and clinical outcome data, widely used for survival analysis and biomarker discovery in oncology. TCGA encompasses multi-omics data from multiple cancer types, while METABRIC focuses specifically on breast cancer and includes high-quality molecular subtyping and long-term survival annotations. In contrast, the TCR--epitope dataset captures adaptive immune recognition by pairing T-cell receptor (TCR) sequences with known antigenic epitopes, forming a sequence-level classification task with applications in immunotherapy and vaccine design. Input features include log-normalized TPM gene expression for genomics datasets and amino acid sequences for the TCR task. Additionally, we evaluate our architecture on a standard computer vision benchmark, CIFAR-10, which comprises 60,000 color images (32$\times$32 pixels) from 10 object categories, split into 50,000 training and 10,000 test samples. Across all datasets, we compare our $L_1$-based transformer to the standard scaled dot-product attention transformer as a baseline.

We frame each task as a multi-class classification problem: predicting cancer subtypes for TCGA and METABRIC, epitope classes from TCR sequences, and object classes for CIFAR-10. All models are trained using cross-entropy loss and evaluated on consistent data splits. To ensure fair comparison, we use identical settings across experiments, including batch size, learning rate, and number of training epochs. For evaluation, we adopt standard metrics for multi-class classification, including overall accuracy, macro-averaged precision, recall, F1 score, and area under the ROC curve (AUROC). As shown in Table~\ref{biovision}, when the $L_1$-based transformer is tuned with an appropriate $\lambda$ value, it consistently outperforms the dot-product-based transformer across all datasets. The performance gains are observed across all metrics: improvements range from $0.01$ to $0.25$ in precision, $0.01$ to $0.06$ in recall, $0.01$ to $0.08$ in F1 score, $0.01$ to $0.05$ in accuracy, and from $-0.02$ to $0.05$ in AUROC.

\begin{table}[h!]
\centering
\caption{Comparison of $L_1$ Attention and Dot-product Attention on Biological and Vision Datasets}
\label{biovision}
\begin{tabular}{llccccc}
\toprule
Data &\textbf{Method} & precision & recall  & F1  & accuracy &AUROC \\
\midrule
TCGA& dot-product  & 0.9834 & 0.9742 & 0.9781 & 0.9814 &0.9387\\
&$L_1$, $\lambda=1$ & 1.0000 & 1.0000 & 1.0000 & 1.0000 &0.9899\\
& $\Delta$ &0.0166 &0.0258 &0.0219 &0.0186 &0.0512\\
\midrule
METABRIC&dot-product  & 0.5869 & 0.5543 & 0.5576 & 0.7500 & 0.8253\\
&$L_1$, $\lambda=10$  & 0.7093 & 0.6194 & 0.6388 & 0.8050 & 0.8045\\
& $\Delta$ &0.1224  &0.0651  &0.0812  &0.0550 &-0.0208\\
\midrule
VDJdb &dot-product  & 0.3058 & 0.2985 & 0.2781 & 0.5066&0.7022\\
&$L_1$, $\lambda=25$  & 0.5656 & 0.3415 & 0.3474 & 0.5390 &0.7067\\
& $\Delta$ &0.2598 &0.0430 &0.0693 &0.0324 &0.0045\\
\midrule
CIFAR-10&dot-product  & 0.7621  & 0.7626 & 0.7623& 0.7626&0.9598 \\
&$L_1,\lambda=1$  & 0.7743  & 0.7732 &0.7736 &0.7732& 0.9626\\
& $\Delta$  &0.0122& 0.0106 &0.0113 &0.0106 &0.0028\\
\bottomrule
\end{tabular}
\end{table}

\section{Conclusion}
We propose an attention mechanism for Transformer models based on the $L1$ distance. Unlike the standard scaled dot-product attention, the proposed method eliminates matrix multiplications in attention score calculation, relying instead on addition and absolute difference operations, which are computationally less expensive. This formulation not only provides a more efficient alternative for attention computation but also offers a flexible framework wherein the attention function can be defined and fine-tuned as a function of distance.

While Transformer models have demonstrated remarkable success across a wide range of applications, their substantial computational and energy demands pose challenges for sustainability. Training and deploying large-scale models require significant electricity consumption and contribute to considerable carbon emissions, especially when scaled across data centers. As AI technologies become more pervasive, improving their energy efficiency is increasingly critical. 
The proposed \textit{EcoTransformer}, with its multiplication-free attention score mechanism, presents a more environmentally responsible architecture, offering potential benefit in energy consumption.





\bibliographystyle{apalike}
\bibliography{EcoTransformerArxiv}
\end{document}